\documentclass[sigconf,screen]{acm/acmart}

\AtBeginDocument{%
  \providecommand\BibTeX{{%
    \normalfont B\kern-0.5em{\scshape i\kern-0.25em b}\kern-0.8em\TeX}}}

\setcopyright{none}
\usepackage{tikz}
\usetikzlibrary{er,arrows,positioning,fit}
\tikzset{
    >=stealth',
    punkt/.style={
           rectangle,
           rounded corners,
           draw=black, very thick,
           text width=6.5em,
           minimum height=2em,
           text centered},
    pil/.style={
           ->,
           thick,
           shorten <=2pt,
           shorten >=2pt,}
}

\begin{document}

\title{Concepts and Paradigms for Neuromorphic Programming}

% \affiliation{%

\author{Steven Abreu}
\affiliation{
  \institution{University of Groningen}
  \city{Groningen}
  \country{Netherlands}}
%\affiliation{
%  \institution{Institute of Neuroinformatics}
%  \city{Zurich}
%  \country{Switzerland}}
\orcid{0000-0002-2272-315X}
\email{s.abreu@rug.nl}


\begin{abstract}
% motivation - programming crucial, current methods lacking, future outlook promising
The value of neuromorphic computers depends crucially on our ability to program them for relevant tasks. Currently, neuromorphic computers are mostly limited to machine learning methods adapted from deep learning.
However, neuromorphic computers have potential far beyond deep learning if we can only make use of their computational properties to harness their full power. 
% claim: paradigm shift necessary
Neuromorphic programming will necessarily be different from conventional programming, requiring a paradigm shift in how we think about programming in general.
% contribution: 1) conceptual cleanup, 2) exploration of paradigms/approaches
The contributions of this paper are 1) a conceptual analysis of what `programming' means in the context of neuromorphic computers and 2) an exploration of existing programming paradigms that are promising yet overlooked in neuromorphic computing. 
% goal: expand horizon
The goal is to expand the horizon of neuromorphic programming methods, thereby allowing researchers to move beyond the shackles of current methods and explore novel directions.

%The value of neuromorphic computers depends crucially on our ability to program them for relevant tasks. Currently, neuromorphic computers are mostly limited to machine learning methods adapted from deep learning. However, neuromorphic computers have potential far beyond deep learning, if we can only make use of their computational properties to harness their full power. Neuromorphic programming will necessarily be different from conventional programming, requiring a paradigm shift in how we think about programming in general. The contributions of this paper are 1) a conceptual analysis of what `programming' means in the context of neuromorphic computers and 2) an exploration of existing programming paradigms that are promising yet overlooked in neuromorphic computing. The goal is to expand the horizon of neuromorphic programming methods, thereby allowing researchers to move beyond the shackles of current methods and explore novel directions.

%Based on existing work from computer and computational science, we explore what neuromorphic programming may look like and what paradigms may be promising, in order to expand the horizon of neuromorphic programming methods.
%, what sort of abstractions may be possible, 
%We embrace the diversity of programming paradigms that may be useful for neuromorphic computing. 
%The ﬁeld of neuromorphic computing is thriving, with many physical realizations at different levels of maturity. 
% configuration is more low-level than programming "programming without abstractions"
\end{abstract}

%\begin{CCSXML}
%<ccs2012>
%   <concept>
%       <concept_id>10010583.10010786.10010792.10010798</concept_id>
%       <concept_desc>Hardware~Neural systems</concept_desc>
%       <concept_significance>500</concept_significance>
%   </concept>
%   <concept>
%       <concept_id>10011007.10011006</concept_id>
%       <concept_desc>Software and its engineering~Software notations and tools</concept_desc>
%       <concept_significance>500</concept_significance>
%       </concept>
%\end{CCSXML}
% 
%\ccsdesc[500]{Hardware~Neural systems}
%\ccsdesc[500]{Software and its engineering~Software notations and tools}

%\keywords{Neuromorphic computing, neuromorphic programming}

\maketitle

% THREADS
% - DIVERSITY of approaches: hardware and programming paradigms (many different tools)
% - beyond digital: mixed-signal and analog
% - computing with physics
% - widen the _sense_ of 'programming'

\section{Introduction}

Computing technology is steering toward impasses, with Dennard scaling ending and Moore's law slowing down \cite{Waldrop2016}. These impasses give rise to innovation opportunities for specialized hardware in computer architecture \cite{HennessyPatterson2019,DeanEtAl2018} as well as in software \cite{Edwards2021}. 
This `Golden Age' of innovation has lead many researchers to investigate neuromorphic computers. Taking inspiration from how the brain computes has a rich history going back at least six decades \cite{Turing1948,Neumann1958} and the recent success of deep learning has demonstrated the power of neural information processing convincingly \cite{LeCunEtAl2015}. 
The development of event-based sensors \cite{GallegoEtAl2022Event,ChanEtAl2007AER}, large-scale neuromorphic processors \cite{Schuman2017}, and brain-computer interfaces \cite{FlesherEtAl2021brain} indicates that neuromorphic computers will play an important role in the future of computing.

An increased diversity of specialized hardware can revive old research ideas or programming paradigms on novel hardware, similar to how the GPU revived research on neural networks for machine learning \cite{Hooker2020,LeCunEtAl2015}.
In light of novel neuromorphic hardware, it is worth re-evaluating overlooked programming paradigms \cite{BanatreEtAl2005}.

Neuromorphic computers take inspiration from the brain, both in the way that information is processed and in the fact that the physical dynamics of the underlying substrate are exploited for computation \cite{Jaeger2021}. 
Research in neuromorphic computing is diverse and happening on many levels: different materials are investigated for basic components in novel computers \cite{TorrejonEtAl2017,TalsmaEtAl2020Synaptic}, different architectures for assembling these components into a computing system are investigated \cite{RuizEulerEtAl2021Dopant}, different domains are considered to move beyond electronics into optical \cite{PorteEtAl2021} or chemical domains \cite{CucchiEtAl2021}. 

A neuromorphic computer is composed of neurons and synapses which model biological neural networks at some level of detail, and they are often implemented directly in the physics of the device \cite{IndiveriEtAl2011}.
Although artificial neural networks (ANNs) are also often considered neuromorphic, this paper focuses on spiking neural networks (SNNs) because they offer a radically different paradigm for computing (see Section \ref{ss:dimensions}), making them an interesting topic for research on programming methods. 

All this requires new theories to describe the computations in such novel devices, along with new theories and methods of programming that can make these devices useful.
The former has been outlined in a recent review \cite{Jaeger2021} whereas the latter is constrained to an as-yet limited set of neuromorphic algorithms \cite{SchumanEtAl2022Opportunities,Aimone2019}.

In Section \ref{s:conceptual-analysis} of this paper, concepts for a more general way of programming neuromorphic computers are analyzed and clarified. To fully harness the potential of neuromorphic computers, algorithm design is not enough. Ultimately, general programming methods must be developed to enable a large group of `neuromorphic programmers' to harness the power of neuromorphic computers for real-world problems beyond machine learning and research benchmarks \cite{Monroe2014Neuromorphic}.

Neuromorphic computers presently cannot be programmed in ways comparable to the rich programming methods of digital computers with instruction set architectures, high-level programming languages, and compilation hierarchies.
Schuman \textit{et al.} \cite{SchumanEtAl2022Opportunities} argue that progress on neuromorphic programming requires a paradigm shift in how to think about programming. Herein, it is assumed that there may not be a single paradigm for neuromorphic programming, just as there is no single paradigm for conventional programming. Section \ref{s:programming-paradigms} stakes out the landscape of programming paradigms to make this body of knowledge available to the neuromorphic community and to identify promising directions for future research.

\section{Concepts}
\label{s:conceptual-analysis}

\subsection{Dimensions of Computing}
\label{ss:dimensions}
% others: continual operation, information/knowledge representation

The brain works quite differently from a digital computer \cite{Neumann1958}. While these differences make it challenging to use conventional programming methods, they simultaneously provide opportunities for novel programming models that are not supported by conventional computers. 
In the following, key differences between conventional and neuromorphic computers are outlined.

\paragraph{Stochasticity}
Neurons are unreliable and noisy \cite{RollsDeco2010}, with neural spike output changing from trial to trial in identical experiments \cite{Maass2014}. Yet, the brain is able to generate reliable behavior from unreliable components. This has fascinated the research community for over six decades and led to models of computing with probabilistic logic \cite{Neumann1956}, stochastic computing \cite{AlaghiHayes2013} where information is represented and processed in probability distributions, and hyperdimensional computing where high-dimensional random vectors are used for distributed data representation and computation \cite{Kanerva2009,ThomasEtAl2021Theoretical}.

\paragraph{Robustness}
The theory of digital computation can be realized robustly in physics through the strong dynamical robustness of bi-stable switching dynamics. The physics of digital computing is very robust, but the theory is brittle in that a single bit flip can lead to catastrophic failure. 
In contrast, the brain works reliably despite the ongoing death and re-generation of neurons. Natural systems like the brain use robust adaptive procedures to work well in unknown and changing environments \cite{SimonLaird2019}.
Mechanisms that provide the physical and functional robustness that natural systems exhibit are only beginning to be understood \cite{Kitano2004}.
%vv done
%HJ: you could also argue that digital computers exploit the very strong dynamical robustness of bi-stable switching dynamics (i.e. two point attractors). It's this switching stability rather than precision engineering which makes digital computers work so well. The physics of digital comp is very robust. The brittleness of digital comp is not a physical brittleness but the bit correctness trap: most programs only run well if every bit operation is correctly programmed.

\paragraph{Distributedness}
In neuromorphic systems, information representation and processing are distributed spatially and possibly also temporally. This is a classical property of neural networks \cite{Rumelhart1986Parallel} which stands in contrast to the localized information in binary transistor states and the sequential execution of elementary instructions in digital hardware.

\paragraph{Unobservability}
While in digital computers every bit of information can, in principle, be addressed, the same is not true in many neuromorphic systems which can only be configured and observed through a limited interface. This prevents the implementation of algorithms that require information which is simply not accessible in neuromorphic systems.

\paragraph{Physical time}
In many neuromorphic computers time represents itself.
In contrast, classical theories of symbolic computation are decoupled from real physical time and simulated through a discrete global clock signal. 
Such decoupling may not be possible (nor desirable) in neuromorphic computers, thus current theories of computation are unsuited for describing neural computation \cite{Jaeger2021}.

\paragraph{Multi-scale dynamics}
Neuromorphic computers operate on multiple temporal scales with no global synchronization, and are often described at multiple different spatial scales: from local learning rules to neural circuits all the way to the global behavior of the network as a whole. Often, the only way to decide what network-level behavior emerges from a local learning rule is to let the network run. This undecidability of global behavior from local rules may be a fundamental property of physical systems that can act as computers \cite{Wolfram1985}.
The difficulty of reasoning about global behavior from elementary operations is solved in digital computing by designing software systems as decomposable hierarchical structures \cite{Simon1991,Booch2011} but this is not presently possible in neuromorphic programming.
%We lack mathematical tools to capture complex interactions of nonlinear phenomena at different timescales, and we lack programming methods to harness dynamics on multiple timescales. 

\paragraph{Analog}
The merits of analog computation in terms of energy efficiency and inherent parallelism are well-known \cite{Boahen2017,Sarpeshkar1998}.
But analog computing is more sensitive to device mismatch and noise which limits the computational depth (number of operations performed in series) \cite{Neumann1958}.
They may also be susceptible to parameter drift, aging effects, and changes in temperature.
%because signals are not restored at every timestep
%because no two analog chips behave exactly the same

\paragraph{No hardware/software separation}
When programming digital computers, one may neglect physical properties of the underlying hardware. 
In neuromorphic computers, such hardware-agnostic programming is not generally possible, as these devices are designed to exploit their underlying \emph{physical} properties and dynamics.
The connection of physical systems and computers has been investigated for decades in the field of unconventional computing \cite{Adamatzky2018Unconventional}, though a general theory of such computation is still missing \cite{Jaeger2021}.
%We need to move physics to computing and back many times before we can hope to decouple HW and SW and develop them.
%gap between physics and computing: different formalisms, no clear bridge between. must do this bit by bit (see \ref{ss:general-programming}. for analog computers, physics is closer.
%Therefore, computing is often considered abstract and unphysical. 
%Digital computing was driven by the theory of symbolic computation and digital computers were built to satisfy this model, using bi-stable circuit elements with fast switching times. 
%%Digital computing has moved from abstract to physical - digital computers were built to satisfy a given model of computation, namely those that are equivalent to Turing machines. 
%%To implement such computations in physical systems, one requires switchable, bi-stable elements such as the transistor in electronics. 
%Neuromorphic computing lies in between these two fields, in that there are various propositions for physical systems and models of computations while they all take inspiration from the brain.
%While it is certainly useful to think about computers in terms of hardware and software, this divide should not be given too much ontological significance as it is evident that any computation must happen in physical space and time \cite{Deutsch1985,Moor1978}. 

\subsection{Physical Computing}
\label{ss:computer-model}

\begin{figure*}[ht]
	\begin{tikzpicture}[node distance=1cm, auto,]
	\node[punkt] (intent) {Intention};
	\node[punkt, above right=of intent] (spec) {Specification}
		edge[pil, <-, bend right=10] node[auto, above left] {formalization} (intent);
	\node[punkt, below right=of intent] (idea) {Idea}
		edge[pil, <-, bend left=10] node[auto] {thinking} (intent);
	\node[punkt, right=4cm of intent] (prog) {Program}
		edge[pil, <-, bend left=10] node[auto, below right=-0.2cm] {coding} (idea)
		edge[pil, <-, bend right=10] node[auto, above right=-0.2cm] {synthesis} (spec);
		edge[pil, <-, dashed, gray] node[auto, above] {NLP} (intent);
%	\draw[->] (prog) to[in=-30, out=-50, looseness=3.8] node[text=gray, midway, below] {compile} (prog);
%	\draw[line width=0.75pt,dashed] (8,2) -- (8,-2); 

	\node[punkt, text width=5em, right=1cm of prog] (inp) {Input};
	\node[text width=6.5em,text centered,right=of inp] (mach) {Machine};	
	\node[text width=6.5em,text centered,below=0.2cm of mach] (conf) {Configuration};
	\node[draw,very thick,rounded corners,fit=(mach) (conf)] {};
	\draw[line width=0.75pt] (11.4,-0.35) -- (13.5,-0.35);
	\node[punkt, text width=5em, right=of mach] (out) {Output};
	\node[punkt, above=2.5em of mach] (env) {Environment};
	\draw [pil,->, shorten >= 4pt] (inp) -- node [name=in] {} (mach);
	\draw [pil,->, shorten <= 4pt] (mach) -- (out);
	\draw [pil,->, shorten >= 4pt] (out) |- (conf);
	\draw [pil,->, shorten >= 4pt] (in) |- (conf);
	\draw [pil,<->, shorten <= 4pt] (mach) -- (env);
%	\draw [pil,->, shorten >= 4pt] (prog) to[in=10] (conf);
	\node[rectangle,draw,very thick,dashed,minimum width = 8.5cm,minimum height = 3.6cm, anchor=north west] (rect) at (8,1.9) {};
	\draw [pil,->,shorten <= 4pt] (prog) -- (rect);
\end{tikzpicture}
	\caption{Diagram of the programming process, see text for explanation. \textbf{Left}: instantiation of a computer program, adapted from Refs. \cite{Gruenert2017,Primiero2016}. \textbf{Right}: organization of a computer.}
	\label{fig:programming-diagram}
\end{figure*}
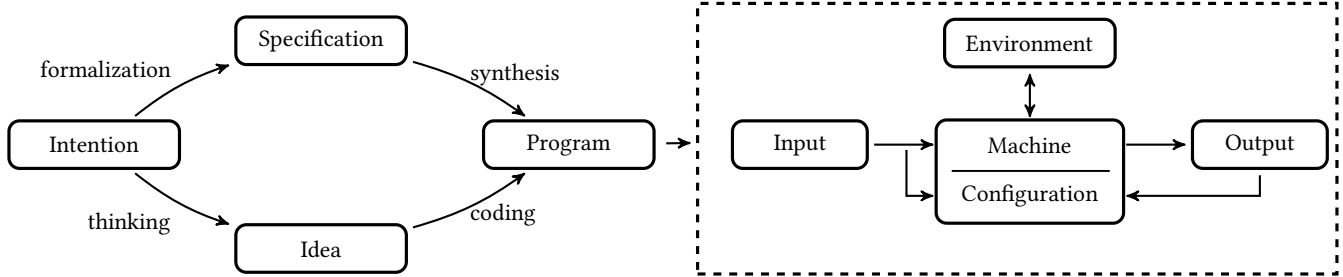

Although classical theories of computing are non-physical, all computations must ultimately be physically instantiated \cite{Deutsch1985}.
Digital computing was first developed as an abstract model which was later physically realized. Neuromorphic computers do not follow the same pattern. There is no universally accepted model of neuromorphic computation and many different physical instantiations are explored \cite{Schuman2017}. As such, abstract models of computation are co-developed with physical implementations. 
From a physical perspective, the key difference between conventional computing and neuromorphic computing lies in the set of physical phenomena that are harnessed for computation. While digital computing only uses bi-stable switching dynamics, neuromorphic computers use stochasticity, real-valued states in continuous time, and more \cite{Jaeger2021}.

Horsman \textit{et al.} \cite{HorsmanEtAl2014} provide a general framework for computation with arbitrary physical systems which was further refined by Jaeger and Catthoor \cite{JaegerCatthoor2022Report}.
Therein, a computer is a physical machine $\mathcal{M}$ which can be stimulated by an input signal $u_\mathcal{M}$ and from which an output signal $y_\mathcal{M}$ can be read out. 
The computation $\mathcal{C}$ is specified by an abstract function from input $u$ to output $y$.
The machine $\mathcal{M}$ then implements the computation $\mathcal{C}$ if an encoding procedure $E$ and decoding procedure $D$ is known such that the machine $\mathcal{M}$ will produce $y_\mathcal{M}$ with $D(y_\mathcal{M}) \approx y$ when stimulated with the input signal $E(u)=u_\mathcal{M}$. 
This leads to the general form of the abstract computer model shown in Figure \ref{fig:programming-diagram} (right): the physical machine $\mathcal{M}$ receives input $u_\mathcal{M}$ and produces output $y_\mathcal{M}$, thereby implementing the abstract computation $\mathcal{C}$ from input $u$ to output $y$. 

\paragraph{Hardware and Software}
Using physics for computation in neuromorphic computers makes it difficult to separate hardware and software in the same way as in digital computers. 
This separation is practically useful because hardware and software are developed on different timescales; it takes many months to design and manufacture a computer chip, while algorithms can be designed and tested within a single day. 
% hardware
Hardware is generally considered to be anything that cannot be changed without significant effort, such as the numbers and types of physical components in the computer. The set of all possible computations that a machine $\mathcal{M}$ can implement is fixed by the hardware.
Considering the hardware to be fixed provides a programmer with firm, stable ground whereon rich, complex programs can be built.
% NOTE: argue contraints of hardware make programming easier?
% software
Software denotes malleable behavioral aspects of the computation $\mathcal{C}$ implemented by the machine $\mathcal{M}$.
Obviously, this behavior is ultimately manifested in the physical state and dynamics of the machine, but it is useful to think of the machine's behavior at an abstract level \cite{Moor1978}. 

\paragraph{Configuration}
In reconfigurable hardware, one must consider the role of a machine's configuration. A reconfiguration of the computer usually requires a reset, effectively breaking the operation of the program. Thus, \emph{the computer's configuration is fixed over the lifetime of a program, but not necessarily fixed over the lifetime of the computer}. 
The configuration can be considered part of the hardware, whereby changing it effectively instantiates a different physical system. But it can also be considered part of the software, whereby changing it simply runs a different program on the same physical system. The chosen view is a design decision by the programmer.

\subsection{Computations and Programs}

%{\color{red}
%would be good to also clarify the relation between the (closely related, apparently) concepts of "program" (according to your view) and "algorithm" (according to the digital conception as "algorithm = Turing machine" for the theoretical view, "algorithm = program written in a programming language" for the practical view).
%}

A computation $\mathcal{C}$ specifies \emph{what} is being computed while a program $\mathcal{P}$ specifies \emph{how} the computation is implemented. There may be many different programs $\mathcal{P}_1,\ldots,\mathcal{P}_n$ that implement the same computation $\mathcal{C}$. 
As such, the computation gives a specification of what is being computed while the program gives a recipe, or mechanism, for how this computation is implemented. 
It is noted that the concept of a `program' herein includes algorithms as Turing machines as well as programs that learn \cite{Valiant2013} and interactive programs, both of which cannot be implemented by Turing machines \cite{Wegner1997,LeeuwenWiedermann2001a}.

In classical computing, a function on natural numbers is implemented by a program which can be represented by a Turing machine.
%This can also be expressed as a function on symbols strings (words), by encoding these symbol strings into numbers. 
In neuromorphic computing, functions that operate on (real-valued) time series are computed. The computation is implemented by a program represented as a neural network, often with designated input and output neurons. 

A computation $\mathcal{C}$ is described by a formal specification which specifies the function that is being implemented. The specification formalizes the informal intention of the computation (Figure \ref{fig:programming-diagram}, left). The specification of a computation is expressed in some mathematical formalism. In digital computing, this can be done using formalisms from logic. In analog computing, there are a variety of formalisms that describe the computation, for example qualitative geometrical constructs like attractors and bifurcations \cite{Jaeger2021}.
%A specification is said to be complete if exactly one function satisfies the specification, and incomplete if more than one function satisfy the specification. 

A program $\mathcal{P}$ is described in another formalism. In digital computing, programs are expressed in some programming language, see Section \ref{ss:languages-paradigms}. In analog computing, one typically uses differential equations to describe the program. When programs interact with another, one may also speak of each individual program as a \emph{process} and the ensemble of all processes as the program, whose behavior emerges from the interaction of the interacting processes (see Section \ref{ss:conventional-programming} on distributed programming).

Operationally, a program is defined by the data flow and control flow.
% data flow and control flow
The data flow specifies how signals that carry computationally relevant information are propagated through the machine. 
The control flow specifies what operations or transformations are done on these signals. 
% examples
For example, in a field-programmable gate array (FPGA) the data flow is configured through its routing elements while the control flow is defined by the function implemented in each logic block. 
In a CPU, data flows between registers and memory according to the program's data instructions while the control flow is defined by its logic instructions.
In a neuromorphic chip, the data flow is defined by the connectivity of the neural network while the control flow is defined by the synapse and neuron models, learning rules, synaptic weights, time constants, thresholds, and more.

\subsection{Programming}
\label{ss:programming}

%Having defined an abstract computer model, and clarified the domain of operation for programmers, this section discusses the programming process itself.
%For present purposes, it is desirable to differentiate between the hardware design of a computer, which is considered fixed, and the programming of the computer's software, which can be done rapidly and repeatedly. 
%As discussed previously, configuration of the physical system can be counted toward the hardware design or the software programming.

Treating hardware as fixed and software as malleable helps to separate the different timescales on which hardware and software are designed. Programming is a software matter and therefore assumes that a physical system already exists which allows to be programmed or configured. 
This does not, however, prevent the programmer from thinking about what properties a physical system \emph{should} have in order to be effectively programmable for some task. On the contrary, this separation of concerns allows clear communication of what hardware constraints 
(there will be constraints!) 
are more or less desirable from a programming perspective, thereby contributing to successful hardware-software co-design.

It has already been mentioned that the physical computing machine is \emph{designed} and \emph{configured} before it can be \emph{programmed}. In the following, some processes which have been called `programming' are delineated, their meanings clarified and a general programming framework is outlined.

\paragraph{Designing} Every computing machine must be designed and manufactured before it can be used. Such machines can be programmable to varying extents. An application-specific computer is not programmable in any way - it physically implements a single program. A reconfigurable computer is configurable but may not be extensibly programmable. A programmable computer is fully programmable. The difference between the latter two depends on their usage and a clear \textit{a priori} separation may not be possible. 

\paragraph{Configuring} Many computing machines can be configured in a way that was defined in the design of the computing machine. 
A configuration can modify the interconnects of an FPGA \cite{KochEtAl2016FPGAs}, the time constants and gains in a spiking neuromorphic chip \cite{MoradiEtAl2018}, or the tunable beam couplers in a photonic circuit \cite{BogaertsEtAl2020}.
As defined in Section \ref{ss:computer-model}, the configuration is constant for the lifetime of the program. 
Configuring is close to the hardware and amounts to selecting a configuration from a pre-defined set of configurations that were designed into the hardware, and is analogous to setting the control systems in a dynamical system.
%Assuming finite precision, the number of configurable values is typically finite, and therefore the set of all possible configurations is also finite (assuming bounded precision).
%{\color{red}
%I am not sure about this in analog systems where you can set real-valued values to memristors, for example. Only after committing to the conception that under finite precision it only makes sense to think of a finite set of reliably distinguishable values one arrives at a finite configuration space. But this space can be combinatorially large (think of configuring 1000 memristors to either high or low resistance: combination space is $2^{1000}$ sized.)
%}
This limits the expressivity and creativity of a programmer constrained to configuration space. 
The configuring is often done through trial-and-error, or automatically through a search procedure if a well-defined objective exists.

% configuring only changes the dataflow once and then its fixed? programming is dynamic and gives instructions also for the future? 
\paragraph{Programming} As opposed to configuring, programming is not strictly constrained by the machine's physical design. The set of all possible programs is typically infinite, providing programmers with an unbounded creative medium for realizing their ideas. 
This infinitude originates in the compositionality of programs.
Moreover, programs have a temporal component; while a configuration is fixed for the program's entire lifetime, a program can change its behavior over time.
%Similar to defining its configurability, the machine's design also has to define its programmability through some interface. 
The key to this expressivity is a programming language in which programs are expressed (see Section \ref{ss:languages-paradigms}).

\paragraph{Optimizing / Training / Learning} 
Programs need not be written manually, but can also be searched for automatically.
Such a search often has a desired program and can therefore be viewed as an optimization problem in which the `distance' between the implemented program and the desired program is minimized. 
The optimization can be done on-device or off-device with a (digital) computer and it can be done either offline in a training phase when the computer is not being used or online while the computer is being used. 
%Generally, interfacing a neuromorphic computing device with a digital computer is costly, 

Training and learning often use optimization methods. In neuromorphic computing, one can \emph{train} a neural network to approximate a desired program through some optimization procedure. A neural network can also \emph{learn} autonomously how to adapt its weights to achieve some global objective, in a self-supervised or unsupervised way. Or it can simply mechanistically apply a learning rule with no clear global objective, like cellular automata \cite{Sarkar2000brief}. Furthermore, using evolutionary algorithms, one may \emph{evolve} a neural network, or a neuromorphic device, to implement some computation.
These approaches are further detailed in Sections \ref{ss:conventional-programming} and \ref{ss:neuromorphic-programming}.

\paragraph{Instructing}
An existing learning algorithm can be further `programmed' through curated interactions with the environment or the user. 
This interactive training is common for personalized AI systems. For example, every Twitter user has a personalized Twitter feed which is learned by the user's behavior but can also be explicitly shaped by hiding or liking certain content. 

\paragraph{Self-organization} 
A popular paradigm for on-chip learning is self-organization. 
Local learning and adaptation mechanisms lead the neural network to self-organize and thereby implement a desired computation, for example with self-organized maps or plasticity rules in SNNs \cite{KhacefEtAl2019Self}. 
As is common with multi-scale dynamics (Section \ref{ss:dimensions}), it may be undecidable which local rules yield a particular global behavior.
Thus, programming with self-organization can be exploratory to investigate what behavior emerges from different local rules, or it can be goal-driven when local rules are optimized to satisfy some behavioral constraints.
Self-organization can also take place directly in physics to grow computing devices where the device is not explicitly designed \cite{CucchiEtAl2021,TalsmaEtAl2020Synaptic,GrzybowskiEtAl2009,BoseEtAl2015}.\\

% HJ (to references): yes! they would be important and very much add substance. And since this article is about "programming", this listing would benefit from being explained in more detail, e.g. in a similar format as your listing of challenges in section 2.  

Figure \ref{fig:programming-diagram} illustrates the general process of programming. Programming begins with some informal intention of what computation the program should implement. This intention can be formalized into a specification, or the programmer may directly come up with an idea for a program that implements the intended computation, expressed in some formalism. This program is then communicated to the physical computer through a pre-defined programming interface. Finally, the system executing this program can be controlled or instructed to remain within the specification.
% program need not be fully known/formalized. Can be just an incomplete idea. The idea can be completed by blanks which are learned/trained/evolved/searched. 

\subsection{Languages and Paradigms}
\label{ss:languages-paradigms}

Conventionally, programming amounts to \emph{coding} (writing source code) in some formal language. Herein, `programming language' is used in an unconventionally wide sense to include any formal language that can be communicated to a physical system. 
This includes programming languages like Python but also extends to other formalisms like differential equations describing dynamical systems, or block diagrams describing signal processing systems. 
In any case, the `programming language' must be compatible with the elementary instructions that the computer's programming interface provides. Given this compatibility, the programmer is free to explore the infinite space of programs.
Work on elementary instruction sets for non-digital computers goes back at least to the 1940s and continues to the present day \cite{Shannon1941,Moore1996,Hasler2020a} but there is still no universally accepted model \cite{JaegerCatthoor2022Report}. 
Consequently, it is not clear what a neuromorphic programming language may look like \cite{MichelEtAl2006}; will it require new syntax such as visual representations, or will a program be represented by a string of symbols in some formal language?
Since the goal is to improve the ``general practice of programming'' neuromorphic computers, Floyd \cite{Floyd1979} argued that it is more effective to turn to programming paradigms rather than to languages. 
A programming paradigm is an approach to programming ``based on a mathematical theory or a coherent set of principles'' \cite{Roy2009} and a programming language implements one or more programming paradigms.
Centering the discussion on programming paradigms shifts the focus away from syntactical issues to the way programs are conceived and designed.

\subsection{Programming Trends}
\label{ss:trends}
% other stuff to mention: design patterns, philosophy/principles (DRY, etc),
% language and program can evolve together -> define new operators

Beyond languages and paradigms, there is a set of well-developed tools for computer programming without which most modern software systems would not exist. Such tools will be necessary if neuromorphic programming is to develop into a mature discipline. 

\paragraph{Efficiency}
Modern programming is done on computers running a powerful integrated development environment (IDE). This is essentially an \emph{interface} between the computer and the programmer, which enables a fast feedback loop between designing and testing programs. The keyboard-and-mouse interface to modern computers now seems trivial, but its success confirms its efficiency.
Programming is interactive, with many intermittent compilation runs to check the program's semantics and syntax, where the syntax is often checked by the IDE directly without compiling. 
%Long gone are the days where programs were submitted to an operator on punched cards, and the results were reported on the next day.

\paragraph{Teamwork}
Much has been invested into the coordination of large software teams \cite{Brooks1995}, resulting in some of the most complex and valuable computing systems in the world \cite{Booch2011}. Collaborative version control systems are used by corporations, organizations, and open-source communities alike, enabling collaboration on large codebases with multiple programmers working on overlapping parts. Agile development and management are commonly used to efficiently coordinate large software projects \cite{DingsoeyrEtAl2012decade}.

\paragraph{Automation}
High-level programming languages elevate the level of abstraction and automate much of the work that was previously done explicitly \cite{Parnas1985}. Furthermore, automated programming techniques are in full force, with inductive programming and machine learning leading the way toward programs that are automatically generated from data (see Section \ref{ss:conventional-programming}). 
%Such data-intensive systems come with new kinds of challenges \cite{SculleyEtAl2015}.

\paragraph{Robustness}
As software systems increase in complexity, much work has been invested to make them robust to failures. Automated testing, continuous integration, and containerization all contribute to making large-scale software development more robust to different kinds of failures \cite{BeyerEtAl2016Site}.
Modularization and structured programming have been key to managing large, interactive, distributed software systems.
But despite significant advances in programming tools, software complexity remains an obstactle for achieving robust systems with no silver bullet in sight \cite{Brooks1987,Potts1993,Larus2009,Ebert2018}.
%However, the essential complexity of the system remains a challenge and makes understanding and maintaining such systems notoriously difficult. 

\paragraph{Software engineering}
Everything above has focused on only one aspect of programming, namely the design of programs. Software engineering can be thought of as ``programming integrated over time'' \cite{WintersEtAl2020} in that it goes beyond the design of programs to also include maintenance, testing, validation, integration, and organization of large software-intensive systems \cite{Booch2011}.

\section{Programming Paradigms}
\label{s:programming-paradigms}

%In the present section, existing programming paradigms are explored. First, conventional programming paradigms are surveyed, with particular attention to paradigms that are relevant for neuromorphic computers. 
%Next, the field of analog programming is outlined, since many neuromorphic computers have analog components. The field of analog programming is less mature, but it is actively being developed \cite{Ulmann2020,Hasler2016}. 
%Finally, a short survey of neuromorphic programming gives an overview of some common approaches to programming neuromorphic computers, and points out some approaches which are yet to be explored.

\subsection{Conventional Programming}
\label{ss:conventional-programming}

\paragraph{Instruction-based}
The most common way of writing sequential, instruction-based programs uses the \textbf{imperative} paradigm, as implemented in C. 
Imperative programming was augmented with objects, which can contain instructions as well as data, to yield the \textbf{object-oriented} paradigm, as implemented in C++ or Java. 

With the advent of multi-core microprocessors came the need to use resources on different cores simultaneously. This led to the development of \textbf{parallel programming} techniques, in which multiple processes are carried out simultaneously on different cores \cite{PachecoMalensek2022Introduction}. 
%Amdahl's law \cite{-Amdahl1967} states that the speedup of parallelization is limited by the ratio of the program that is inherently sequential. 
This is not to be confused with \textbf{concurrent programming} where the lifetime of multiple computing processes overlap and may interact with another \cite{Milner1993}. Concurrency introduces issues of synchronization such as deadlocks and race conditions.
%Petri nets are a general, graph-based formalism which can model many sorts event-driven, concurrent processes, not only parallel unclocked computer programs but also real-world systems like production processes in factories. 
\textbf{Distributed programming} deals with programs that are executed on multiple networked computers which interact to achieve a common goal. 

\textbf{Emergent programming} uses multiple interacting sub-programs whose collective behavior constitutes the desired program. The individual instructions are typically not explicitly informed of the program to be created \cite{GeorgeEtAl2005Experiments}. This approach has been used to design programs that exhibit some creativity \cite{Hofstadter1991}. This is reminiscent of local learning rules in neuromorphic computers (see Section \ref{ss:neuromorphic-programming}).
%, by using methods from adaptive multi-agent systems and relying on cooperative self-organization \cite{GleizesEtAl1999}. 

\paragraph{Declarative}
Instead of describing the control flow of a program, \textbf{declarative} programs describe the logic of the program. A declarative program describes \emph{what} the program does rather than \emph{how} it does it. This makes reasoning about programs easier and simplifies parallel programming \cite{Backus1978}. Declarative programming is done in database query languages like SQL, functional programming languages like Haskell, or logic programming languages like Prolog.
In \textbf{dataflow programming}, a program is modeled as a graph of data flowing between operations. This is a natural model for neuromorphic computers where data flows between neurons, and has been used for neuromorphic compilation \cite{ZhangEtAl2020} (see Section \ref{ss:neuromorphic-programming}). 
\textbf{Spatial programming} can be used to program reconfigurable hardware into dataflow engines \cite{BeckerEtAl2016}.
%This allows for \emph{massively} parallel computation, similar to cellular automata.

\paragraph{Automated programming}
In \textbf{meta-programming}, it is possible for a program to write or modify programs, by simply treating the program as data. 
In \textbf{reflective programming}, a program modifies its own behavior whereas in \textbf{automatic programming}, a program generates another program. 
If a formal specification of the desired program is given, \textbf{program synthesis} can be used to generate a program that provably satisfies this specification \cite{GulwaniEtAl2017}.
If exact adherence to a formal specification is not required, but only the satisfaction of given constraints, \textbf{constraint programming} may be used \cite{RossiEtAl2008Constraint}.
If an incomplete specification is available, such as input-output examples, then \textbf{inductive programming} can be used to generate a suitable candidate program \cite{GulwaniEtAl2015Inductive}.
%The idea of inductive programming may be even further extended to natural language programming (NLP, see Figure \ref{fig:programming-diagram}) where a computer program generates a desired program from an input given in natural language. 
An inductive programming approach coupled with probabilistic programs has been proposed as a model for human-level concept learning \cite{LakeEtAl2015}.
Recently, deep learning (see below) has been used for inductive programming, under the name of neural program synthesis \cite{Kant2018}. 
%end-user programming
As already mentioned in Section \ref{ss:programming}, it is possible to instruct an interactive program and direct it to implement a desired computation. \textbf{End-user programming} allows users to obtain programs from a small set of examples, like the flashfill feature in spreadsheet programs which infers a formula from table manipulations done by the user \cite{Schmid2018,GulwaniEtAl2015Inductive}. 

\paragraph{Probabilistic}
While classical programs are deterministic, the execution of a probabilistic program depends on random numbers, for example by calling a (pseudo) random number generator. Such a program can be viewed as sampling from a probability distribution. In \textbf{probabilistic programming}, the program itself is considered to be a distribution, and the programmer can analyze this distribution and condition the distribution on observations \cite{GordonEtAl2014}. Indeed, the goal of probabilistic programming is not simply the execution of a program, but also the analysis thereof. 
By expressing a statistical model as a probabilistic program, statistical inference on such a model can be done automatically by the compiler through general-purpose inference schemes. 
Probabilistic programming has been used for state-of-the-art generative vision models with very compact programs of only about 50 lines \cite{KulkarniEtAl2015}.
% For example, stochastic computing is about getting a good enough answer even with errors. Fuzzy logic is actually closer to the concept that we’re talking here, where you’re deliberately keeping track of uncertainties as you process information
% further reading: http://adriansampson.net/doc/ppl.html and http://www.pl-enthusiast.net/2014/09/08/probabilistic-programming/

\paragraph{Learning}
%Machine learning, deep learning, differentiable, clarify connection to optimization, probabilistic
In classical programming, a human programmer defines the program that specifies how input data is processed. \textbf{Machine learning} constructs programs that learn from the input data, in ways that may not have been anticipated by any human. Machine learning has deep roots in probability theory and overlaps significantly with probabilistic programming \cite{Murphy2012Machine,Ghahramani2015}.
%For the present context, supervised machine learning and reinforcement learning are particularly relevant. 
In supervised machine learning, a mapping from inputs to outputs is learned from a set of examples. 
In reinforcement learning, a policy of how to act in some environment is learned from rewards and punishments. 
Both the learned mapping in supervised learning and the learned policy in reinforcement learning can be used as programs. This makes machine learning a new paradigm for (automated) programming \cite{CholletAllaire2018Deep}.
%Supervised ML can be used to train a ML model to approximate a function given a set of input-output examples. The ML model can be used as a program, thereby making this a way of programming. 
% Machine learning has already brieﬂy been described, and while a thorough treatment is out of scope for our purposes, we identify the key idea as learning a machine learning model which represents a program, in the form of an input-output mapping from given input-output examples (supervised learning), or a behavior that is learned from some reward signal in an environment (reinforcement learning).
% connection to optimization
Machine learning uses tools from optimization theory; the learning task is often partly framed as an optimization problem where some surrogate of the true performance metric is optimized, for example the average error over a set of input-output examples. 
%With objective, or without objective.
%Uses techniques from optimization, but is still quite different. We optimize on a surrogate loss function, but since this is not the actual performance measure we care about, there are some additional caveats (early stopping etc).
% https://cedar.buffalo.edu/~srihari/CSE676/8.1%20LearningVsOptimizn.pdf
%NOTE: learning in ANNs as a paradigm \cite{Barreto1990}

In \textbf{reservoir computing}, a neural network consists of an input layer which feeds into the high-dimensional recurrently-connected reservoir network from which the output is obtained through a readout layer. Only this final readout layer is trained while the reservoir is randomly initialized and typically not modified (see Section \ref{ss:neuromorphic-programming}).
\textbf{Deep learning} uses multi-layered ANNs for machine learning. The connectivity of such an ANN is usually fixed and then the weights are learned, typically in a supervised fashion using gradient descent to minimize the error on given input-output examples.
In \textbf{differentiable programming}, programs are written in a way that they are fully differentiable with respect to some loss function, thereby allowing the use of gradient-based optimization methods to find better-performing programs. Deep learning is a special case of this, where programs are artificial neural networks that are differentiated using backpropagation. These techniques have also been adapted for spiking neural networks \cite{NeftciEtAl2019}.
Differentiable programming has been employed to merge deep learning with physics engines in robotics \cite{DegraveEtAl2019}, it has been applied to scientific computing \cite{InnesEtAl2019}, and even towards a fully differentiable Neural Turing Machine \cite{GravesEtAl2014}.

\paragraph{Optimization}
As already mentioned, machine learning relies heavily on tools from optimization theory. In pure optimization, the minimization of some cost function $J$ is a goal in itself. 
In machine learning, a core goal is good generalization to unseen examples. This is expressed as some performance measure $P$ which is intractable and therefore one minimizes some cost function $J$ which will in turn also increase the performance measure $P$.
As such, if generalization is not needed then one may use optimization as a programming paradigm in which the result of the optimization is the desired program or the optimization process itself.
\textbf{Evolutionary programming} uses population-based evolutionary optimization algorithms to find programs. In order to find a program that solves some problem is to define a fitness function that is maximized by a program that solves this problem.
Evolutionary algorithms have been used to generate rules for a cellular automaton to solve computational problems that are difficult to solve by manually designing a learning rule \cite{MitchellEtAl1994}. Evolutionary optimization is also a popular approach for neuromorphic devices, see Section \ref{ss:neuromorphic-programming}.\\

Some dimensions of neuromorphic computing (Section \ref{ss:dimensions}) are exploited by paradigms in this section. Dataflow programming, distributed programming and deep learning harness distributedness in computation. Probabilistic programming uses stochasticity, as do optimization methods and machine learning methods. Emergent programming works with at least two different spatiotemporal scales as well as learning and optimization where the optimization loop operates on a slower timescale than the actual program. In some machine learning and optimization methods like reservoir computing or evolutionary optimization, a complete description of the program is not necessary, potentially accommodating some unobservability.

\subsection{Unconventional Programming}
\label{ss:unconventional-programming}

The present section investigates paradigms for programming physical systems.
Computational models, and therefore programming methods, must ultimately be based in physics and resulting hardware constraints \cite{Stepney2012a}. 
Current programming methods are adapted to clocked digital hardware but with the forthcoming diversity of computer hardware and architectures \cite{HennessyPatterson2019} it is time to widen the set of hardware constraints that can be programmed with.

\paragraph{Cellular programming}
As mentioned previously, cellular automata (CA) are a standard model of massively parallel computation. A CA is programmed by choosing its update rule and the program is executed on some initial configuration of the CA's lattice.
Inspired by CAs, cellular architectures of neuromorphic devices have been proposed \cite{KhacefEtAl2020Brain,Schuman2017}.
For over two decades, \textbf{amorphous computing} has been developing programming techniques inspired by the cellular cooperation in biological organisms \cite{Stark2013Amorphous}. An amorphous computer is a system of irregularly placed, asynchronous, locally interacting computing elements that are possibly faulty, sensitive to the environment, and may generate actions \cite{AbelsonEtAl2000,Coore2005Introduction}. This line of research brought space-time programming \cite{BealViroli2015} as a way of programming to control large networks of spatially embedded computers. 
Although not directly focused on neuromorphic computing, amorphous programming methods can provide a good starting point for robust programming methods in cellular architectures.

\paragraph{Analog programming}
Neuromorphic hardware often contain analog components, which are difficult to work with because programming methods for analog computers are not at the same level of maturity as those for digital computers. Ulmann \cite{Ulmann2020} argues that the development of reconfigurable analog computers will advance the state of analog computer programming and efforts to develop such hardware is in progress \cite{Hasler2020}.
Nevertheless, methods from control engineering, signal processing and cybernetics have been developed and used for decades and can be adapted for neuromorphic systems. 
While digital computing was originally formulated as computing functions on the integers \cite{Turing1937}, \textbf{signal processing} can be seen as computing functions on temporal signals. For analog neuromorphic computers, signal processing provides a rich framework for computing with temporal signals \cite{RajendranEtAl2019Low,DonatiEtAl2018Processing}.

\textbf{Control theory} has developed a rich repertoire of methods to drive a dynamical system into a mode of operation that is robust, stable, and implements some desired dynamics. These methods can be used to keep analog computers within a desired regime of operation to implement a desired computation.
It can be expected that analog computers can benefit from cross-fertilization between computer science and control theory \cite{MichelEtAl2006}.
A promising direction is data-driven control where a model of the system to be controlled is learned from experimental data using machine learning techniques \cite{BruntonKutz2019Data}.
Historically rooted in ideas from cybernetics and ultrastable systems \cite{Ashby1960Design}, \textbf{autonomic computing} aims to design systems that are able to adapt themselves in order to stay within a high-level description of desired behavior \cite{ParasharHariri2005}. The field takes inspiration from the autonomic nervous system, which is able to stay within a stable `dynamic equilibrium' without global top-down control. 

\paragraph{Programming physical systems}
%In neuromorphic computers, and other unconventional computers \cite{Adamatzky2018Unconventional}, the computation is directly implemented in the physics which builds a close interplay between physics and computing that is not present in digital computing and which current cannot be fully exploited for a lack of theory \cite{Jaeger2021}. Nevertheless, approaches exist which can harness the dynamics of a physical system for computation. 
Building on evolutionary optimization, \textbf{evolution \textit{in materio}} \cite{MillerDowning2002} was proposed to harness material properties for computation. It is argued that natural evolution excels in exploiting the physical properties of materials, and artificial evolution emulates this. Evolution has been applied widely in unconventional computing \cite{Adamatzky2018Unconventional}, for example with a disordered dopant-atom network for digit classification \cite{ChenEtAl2020}.
As already mentioned in the preceding section, \textbf{physical reservoir computing} can be used to harness the dynamics of physical systems for computation by modeling the physical system as a high-dimensional reservoir on top of which an output map is trained \cite{TanakaEtAl2019}.

\subsection{Neuromorphic Programming}
\label{ss:neuromorphic-programming}
%DC has had decades to build abstraction and compilation hierarchies to facilitate the programming of increasingly complex computer systems.

%Schuman: backprop, conversion, reservoir, evolution, plasticity
%Further: non-ML algs, programming abstractions

%Offline off-device optimization methods are most mature as they have been advanced by computer science and machine learning for decades. 
%
%Off-chip online optimization methods also benefit from extensive work done in the fields of computer science, machine learning, and signal processing.
%{\color{red}TODO give examples}.
%
%{\color{red}TODO rewrite}
%
%Some offline optimization methods can be implemented directly on-device, such as equilibrium propagation \cite{KendallEtAl2020Training}, and efforts exist to implement gradient-based methods on-device as well \cite{BoonEtAl2021Gradient}.
%
%Off-chip online optimization methods also benefit from extensive work done in the fields of computer science, machine learning, and signal processing. {\color{red}TODO give examples}.
%These methods can also be adapted to the on-device setting, although this is often difficult in practice because the computations required to implement these optimization methods are often costly in terms of area and energy on device. For this reason, self-organization methods are a popular alternative for online on-device learning. 
%In an on-device, online settings, optimization methods from the off-device, online setting can be implemented. 

\paragraph{Neuromorphic co-design}
% manual design -> starting point
% gives an upper bound for the flexibility one can have
As neuromorphic computers exploit physical phenomena of their underlying hardware, manually designed neuromorphic programs will necessarily be close to physics. Therefore, although not strictly a paradigm for `programming', it is instructive to consider \textbf{neuromorphic co-design} as a paradigm for designing neuromorphic systems.
The field is rooted in the original vision of neuromorphic computing \cite{Mead1990} and designs application-specific \cite{MastellaChicca2021} as well as reconfigurable \cite{MoradiEtAl2018} mixed-signal neuromorphic chips in sub-threshold CMOS technology which may also include on-chip learning. 
This approach uses tools from signal processing and computational neuroscience to implement a desired behavior in networks of silicon neurons \cite{IndiveriEtAl2011}. 
Similar to analog computing, the field may benefit from a set of computational primitives to simplify the design of neuromorphic systems.

\paragraph{Compilation}
Given a neural network, it is necessary to communicate this network to the hardware. 
\textbf{Neuromorphic compilation} \cite{ZhangEtAl2020} was proposed as a general framework to (approximately) compile neural networks into different hardware systems, automatically adapting to physical constraints. 
Such compilation can be done statically to exactly implement the specified network architecture \cite{GruauEtAl1995neural,Siegelmann1994Neural}, or adaptively to further optimize the network after compilation \cite{BunelEtAl2016Adaptive}. In any case, it is important to consider the hardware constraints in this compilation \cite{JiEtAl2018Bridge}.

To compile a neural network into hardware, it is necessary to first design the neural network's architecture. Deep learning has accumulated a plethora of well-performing network architectures for ANNs which can rapidly be converted into equivalent spiking neural networks (SNNs) through ANN-to-SNN \textbf{conversion} \cite{DiehlEtAl2016Conversion,RueckauerEtAl2017Conversion}. 
The conversion to SNNs offers significant advantages in energy efficiency while often maintaining similar levels of performance. However, this conversion is not optimal because it typically does not leverage the computational power of spiking neurons and instead limits the richer dynamics of SNNs to the same less powerful domain of ANNs \cite{SchumanEtAl2022Opportunities}.

Compilation and conversion are promising directions, though descriptions at the level of neural network architectures may not provide a high enough abstraction for implementing programs that realize arbitrary computations. 

\paragraph{Learning}
Given the success of deep learning, learning is a natural paradigm for neuromorphic computers. 
While it would be naïve to ignore the deep learning literature, it is also unrealistic to expect deep learning methods to work for SNNs as well as they do for ANNs since these methods were optimized for ANNs \cite{DaviesEtAl2021Advancing}. 
%Deep learning uses backpropagation to compute the synaptic weight updates to do gradient descent on some loss measure. 
%Naturally, after compiling a neural network onto neuromorphic hardware, it is often necessary to re-train the neural network. Training and learning in neuromorphic hardware is challenging because of limited observability and other issues mentioned in Section \ref{ss:dimensions}.

\textbf{Backpropagation}, the workhorse of deep learning, can be implemented directly in SNNs using surrogate gradients \cite{NeftciEtAl2019} or other neuromorphic adaptations. Simplifications of the backpropagation algorithm such as the random backpropagation algorithm \cite{BaldiEtAl2018Learning} were also demonstrated in neuromorphic systems \cite{NeftciEtAl2017}. 
It is also possible to create a \textbf{surrogate model} of the physical device, then optimize the surrogate model in simulation with deep learning methods and transfer the optimized model back to the device \cite{EulerEtAl2020deep,HeEtAl2019}.

For recurrent neural networks, \textbf{reservoir computing} avoids the need to backpropagate information through the network to compute gradients. Instead, the reservoir is kept fixed and only a readout map from the reservoir to the output is trained. This training procedure requires the reservoir states to be read out and stored, which may not be possible given limited observability of some devices or limited data storage. Reservoir computing is a popular paradigm for neuromorphic computing, with dedicated frameworks for hardware implementation \cite{SchumanEtAl2022Opportunities,Michaelis2020}.

Neural network training is often done off-device with external hardware. Frequent re-training creates a large overhead, limiting the performance and applicability of neuromorphic computers. As a result, \textbf{on-device learning} methods are an active topic of research \cite{BasuEtAl2018Low}. 
\textbf{Plasticity} is a popular paradigm for on-device learning where local learning rules are used to modify the connectivity (structural plasticity) and connection strengths (synaptic plasticity) of a SNN.
Parallels to emergent programming may be drawn here as the resulting behavior of the SNN emerges from the interaction of local rules. It is not clear what local rules will yield a particular network-level behavior, but evolutionary search \cite{JordanEtAl2020} and meta-learning \cite{ConfavreuxEtAl2020} have been used to (re-)discover desirable plasticity rules.
%Plasticity methods have been successfully used for unsupervised learning \cite{DiehlCook2015Unsupervised}, clustering, Bayesian{\color{red}... (refs)}.
% clustering, bayesian inference

\paragraph{Evolution}
A key advantage of evolutionary approaches is that they can jointly optimize the network's architecture and weights, thus simultaneously designing and training the network. Moreover, evolutionary methods do not require differentiability of activation functions, nor do they place any constraints on the network's architecture. 
Evolutionary approaches can find a SNN by randomly choosing an initial population of candidate SNNs, selecting the highest-performing candidates according to some performance metric, and then creating new candidates through recombining and mutating the selected candidates \cite{SchumanEtAl2020Evolutionary,SchliebsKasabov2013Evolving}. 
However, evolutionary approaches can be slower to converge than other training methods \cite{SchumanEtAl2022Opportunities} and the resulting architectures are not easily understandable or reusable for different tasks \cite{Conrad1988}.
%their outcomes do not usually translate to different tasks - a neural network for facial recognition optimized by evolutionary methods cannot easily be reused for another computer vision task. 
%{\color{red}explain this last point better}

\paragraph{Neuromorphic algorithms}
With the increased availability of neuromorphic hardware, a number of handcrafted spiking neuromorphic algorithms (SNA) have been proposed. SNAs implement computations using temporal information processing with spikes, often to implement well-defined computations such as functions on sets of numbers \cite{VerziEtAl2018Computing}, functions on graphs \cite{HamiltonEtAl2019Spike}, solving constraint satisfaction problems or solving a steady-state partial differential equation using random walks \cite{SmithEtAl2020Solving}.
SNAs are being actively developed and many application domains are yet to be explored \cite{Aimone2019}.
% A number of handcrafted SNN algorithms have been proposed in recent years to solve well-deﬁned computational problems using spike-based temporal information processing. When implemented on neuromorphic architectures, these algorithms promise speed and efﬁciency gains by exploiting ﬁne-grain parallelism and event-based computation. Examples include computational primitives, such as sorting, max, min, and median operations [70], a wide range of graph algorithms [71]–[74], NP-complete/hard problems, such as constraint satisfaction [75], boolean satisﬁability [76], dynamic programming [77], and quadratic unconstrained binary optimization [78], [79], and novel Turing-complete computational frameworks, such as Stick [80] and SN P [81].

% primitives and synthesis
\paragraph{Neurocomputational primitives}
A variety of neurocomputational primitives have been proposed in the neuromorphic community. Such primitives can be useful for simple tasks and typically allow for composability to create more complex neuromorphic systems at a higher level of abstraction \cite{BartolozziEtAl2022Embodied,MarcusEtAl2014}.
% WTA & DNF
\textbf{Dynamic neural fields} (DNFs) are a modern framework for neural attractor networks \cite{Schoener2019}. The stable states provided by attractor dynamics help with the intrinsic variability of analog neuromorphic circuits and have been shown to be a promising abstraction for neuromorphic programming \cite{DaviesEtAl2021Advancing}.
Each DNF is a network of neurons that is, under some constraints, computationally equivalent to a \textbf{winner-take-all} (WTA) network \cite{Sandamirskaya2014Dynamic}. The WTA is a common circuit motive in the neocortex \cite{DouglasMartin2007Recurrent}.
%Computation with attractor networks has a rich history in neural computation \cite{Hopfield1982} and provides a simple way for ....... neurosymbolic integration.
% NSM
The \textbf{neural state machine} (NSM) \cite{NeftciEtAl2013,LiangEtAl2019Neural} also builds on WTA networks to implement finite state machines in SNNs, and has been shown to run robustly on mixed-signal neuromorphic hardware.
% sPLL
The \textbf{spiking phase-locked loop} (sPLL) \cite{MastellaChicca2021} was designed for frequency detection as part of a neuromorphic tactile sensor.
% TDE
The \textbf{temporal difference encoder} (TDE) \cite{MildeEtAl2018,GutierrezGalanEtAl2021Event} is a spiking model that was designed to compute the time difference between two consecutive input spikes. The number of output spikes and the time between them is inversely proportional to the time difference. This has been used for motion estimation and obstacle avoidance \cite{MildeEtAl2017Obstacle}.
%Delay/Temporal measurement circuits take inspiration from the insect brain, where motion is computed as the time to travel of a stimulus from one sensing element to the neighbour 180 . This type of computational primitive is useful for motion estimation and obstacle avoidance 88 
% for temporal encoding on event-based signals, which translates the time difference between two consecutive input events into a burst of output events. The number of output events along with the time between them encodes the temporal information.
%The TDE unit translates the time difference between two events into a burst of output spikes. Both the number of output spikes and the duration of the burst produced by the model directly reflect the temporal correlation of two input signals, and it is inversely proportional to the time difference
\textbf{Neural oscillators} generate rhythmic activity that can be used for feature binding and motor coordination, for example as a central pattern generator \cite{KrauseEtAl2021Robust}.
% Further
Other primitives are scattered around the literature and shared libraries of neurocomputational primitives are only starting to be assembled \cite{BartolozziEtAl2022Embodied}.
% into a library of neurocomputational primitives which can then be used to build increasingly complex operations \cite{BartolozziEtAl2022Embodied,MarcusEtAl2014}
%delay measurements \cite{MildeEtAl2017Obstacle,BartolozziEtAl2022Embodied},
% cooperative-competitive networks: e.g. Relational networks use recurrent connectivity to express relative dependencies between variables, for example to compute the error between a measured signal and its target value 78 .
%, or relational networks \cite{DiehlEtAl2018Factorized}.
%, and more \cite{BartolozziEtAl2022Embodied}.
%\cite{CarandiniHeeger2011Normalization} normalization
\textbf{Neuromorphic synthesis} \cite{NeftciEtAl2013} may provide a systematic way of programming complex high-level behavior into neuromorphic chips. This was demonstrated for functions that can be described by finite state machines, but it may be promising to extend this work to a larger set of computational primitives for higher abstractions in neuromorphic programming.

\paragraph{Higher abstractions}
% NEF
The \textbf{neural engineering framework} \cite{EliasmithAnderson2002Neural} raises the level of abstraction beyond the level of neural networks. This allows dynamical systems to be distilled automatically into networks of spiking neurons that can then be compiled down to mixed-signal spiking neuromorphic accelerators like Braindrop \cite{NeckarEtAl2019} using the Nengo programming environment \cite{BekolayEtAl2014}.
% VSA
%The \textbf{vector symbolic architectures} (VSA) framework is suitable for neuromorphic systems, enabling knowledge representation and reasoning in high-dimensional spaces, building on the same principles as hyperdimensional computing (see Section \ref{ss:dimensions}).
%Vector symbolic architectures (VSAs) offer a mathematical, connectionist framework that supports rich knowledge representations and reasoning in high-dimensional spaces [135]. VSAs interfaced with deep networks and generalizations of the optimization and search algorithms described in this survey could provide a path to enabling fast, efﬁcient, and scalable next-generation AI capabilities on neuromorphic hardware.

% Lava
Intel recently launched \textbf{Lava}\footnote{\url{https://github.com/lava-nc/lava}}, an open-source neuromorphic programming framework for the Loihi chips. Support for other neuromorphic hardware is planned. Lava is a multi-paradigm framework and includes libraries of neuromorphic algorithms for optimization, attractor networks, deep learning methods for SNNs, VSAs, and plans to include more paradigms.

% Fugu
Aimone \textit{et al.} \cite{AimoneEtAl2019Composing} proposed \textbf{Fugu}, a hardware-independent mechanism for composing different SNAs. In Fugu, a program is specified as a computational graph reminiscent of dataflow programming, where nodes represent SNAs and connections represent dataflow between the SNAs. 
%Each SNA has a certain input size and output size (number of inputs and time length) in addition to a circuit depth which specifies how many timesteps are necessary for incoming data to be processed. The entire program is specified by the neural circuit that includes all SNAs and their interconnections. 
This program can then be compiled into different hardware-specific configurations. The focus of this work is on digital neuromorphic processors and support for mixed-signal hardware is not discussed.

\section{Outlook}

% computing and physics gap
Without a guiding theory that unites physics with computation, it is difficult to program computers that harness their underlying physical dynamics for computation. Building on decades of research in neuromorphic computing and engineering, initial features of neuromorphic programming methods can be identified.

% Formalizing, making protocols, defining standards
As the field is moving toward general programming methods, it is important to clarify concepts and establish an efficient separation of concerns to allow effective cross-disciplinary collaboration and communication. 
Shared benchmarks and user-friendly tools will further boost progress \cite{BartolozziEtAl2022Embodied}.
Moreover, for neuromorphic systems to scale in the landscape of large heterogeneous computing systems, community-wide standards and protocols must be defined for the communication between neuromorphic systems. 

%% overview / vision of neuromorphic programming
%Several directions have been outlined: 

The structure of large-scale neuromorphic programs is yet to be explored. 
%Jonas and Kording \cite{JonasKording2017} argue that a
It is assumed that a digital computer has a clearer architecture with fewer modules whereas the brain has a larger breadth of ongoing computations \cite{JonasKording2017}.
It remains to be seen if neuromorphic programs allow for the kinds of `crisp abstractions' \cite{Booch2011} that enable the deep hierarchies in digital programming as observed in compilation hierarchies, function call hierarchies, and class inheritance hierarchies. 
If such abstractions are not possible, hierarchies in neuromorphic programs will necessarily be wide and shallow, leading to many interacting components and only a few different levels of abstractions. % - perhaps the neuronal level, the computational primitive level, and the global level.
%{\color{red}FINETUNE}
% can be cleanly decomposable as digital programs are or if they will necessarily be only \emph{nearly} decomposable, as natural systems tend to be \cite{Booch2011,Simon1991}.
%In the latter case, we may be limited to neuromorphic programs whose hierarchy 
%cannot be as deep as those in digital computer programs. 
%is much wider and more shallow than digital computer programs are.
%As such, the breadth of ongoing computation in the processor may actually be simpler than those in the brain.
%Yet many of the differences should make analysing the chip easier than analyzing the brain. For example, it has a clearer architecture and far fewer modules.

%Neuromorphic programming language
%In any case, the `programming language' must be compatible with the elementary instructions that the computer's programming interface provides. Given this compatibility, the programmer is free to explore the infinite space of programs.
%Work on elementary instruction sets for non-digital computers goes back at least to the 1940s and continues to the present day \cite{Shannon1941,Moore1996,Hasler2020a} but there is still no universally accepted model \cite{JaegerCatthoor2022Report}. 
%At this point, it is not clear what a neuromorphic programming language may look like. 

%% sketch out promise / vision for neuromorphic programming in the future
%Promise and vision for neuromorphic programming

% Programming in the large (from simple functions to complex heterogeneous computing systems)
It is hoped that neuromorphic programmers can leverage the work outlined in this paper to build large-scale neuromorphic programs to tackle real-world tasks, and to further develop guiding principles and paradigms for neuromorphic programming. % `in the large' \cite{MichelEtAl2006}.
\begin{acks}
I wish to thank Herbert Jaeger for helpful comments.
This project has received funding from the European Union's Horizon 2020 Research and Innovation Programme under the Marie Skłodowska-Curie grant agreement No. 860360 (POST DIGITAL).
\end{acks}

%\clearpage
\bibliographystyle{acm/ACM-Reference-Format}
\bibliography{cleaned_refs.bib}
%\bibliography{../../Documents/Papers/_references.bib}

\appendix

\end{document}